\documentclass[journal]{IEEEtran}

\usepackage{graphicx}
\usepackage{subfigure}
\usepackage{url}
\usepackage{times}
\usepackage{amssymb}
\usepackage{array}
\usepackage[ruled,lined]{algorithm2e}
\usepackage{amsmath}
\usepackage{booktabs}
\usepackage{multirow}
\usepackage{cite}
\usepackage{enumerate}
\usepackage{pifont}

\begin{document}

\title{Strong but Simple Baseline with Dual-Granularity Triplet Loss for Visible-Thermal Person Re-Identification}

\author{Haijun Liu, Yanxia Chai, Xiaoheng Tan, Dong Li and Xichuan Zhou
\thanks{}
\thanks{H. Liu, Y. Chai, X. Tan, D. Li and X. Zhou are with the School of Microelectronics and Communication Engineering, Chongqing University, Chongqing, 400044, China.}
}

\markboth{}
{Liu \MakeLowercase{\textit{et al.}}: }

\maketitle

\begin{abstract}
This letter presents a conceptually simple and effective dual-granularity triplet loss for visible-thermal person re-identification (VT-ReID). Generally, ReID models are always trained with the sample-based triplet loss and identification loss from the fine granularity level. Further, center-based loss could be introduced to encourage the intra-class compactness and inter-class discrimination from the coarse granularity level. Our proposed dual-granularity triplet loss well organizes the sample-based triplet loss and center-based triplet loss in a hierarchical fine to coarse granularity manner, just with some simple configurations of typical operations, such as pooling and batch normalization.
Experiments on RegDB and SYSU-MM01 datasets show that with only the global features our dual-granularity triplet loss can improve the VT-ReID performance by a significant margin. It can be a strong VT-ReID baseline to boost future research with high quality.

\end{abstract}


\section{Introduction}
\label{sec:intro}

\IEEEPARstart{V}{isible}-thermal person re-identification (VT-ReID), aiming to search a person of interest cross-modality and cross-sensor cameras deployed at different locations, is widely encountered in a practical 24-hour intelligent surveillance scenarios, especially during the nighttime \cite{tifs19vtreid}. Compared to traditional visible-visible ReID (VV-ReID), which focuses on retrieving pedestrians only cross RGB cameras \cite{Ye2020DeepLF,Liu2019Gallery}, VT-ReID is a more challenge problem, since person images are from different modalities with a huge gap. Apart from the intra-modality (intra- and inter-class) variations as existed in VV-ReID, VT-ReID additionally suffers from the large cross-modality discrepancy.

Recently, due to the broad application prospects of VT-ReID, an large number of studies with some novel and effective modules or training strategies are presented to address this problem \cite{liu2020enhancing,ye2020dynamic}. However, many works ignored the design of baseline model, just evaluating the effectiveness of their ideas with a poor baseline. It has negative effect on the developing of VT-ReID community, since the improvement of baseline model plays an important role. \textbf{Therefore, the present study focuses on developing a strong and effective VT-ReID baseline with some simple and typical means.}

Generally, to simultaneously address the intra-modality variations and cross-modality discrepancy, different methods have been proposed, mainly focusing on model designing and metric learning.
Moreover, in order to obtain great results, researchers in the academia always aggregate several part (local) features \cite{Wang2020MPMN} or leverage semantic features from pose estimation \cite{su2017pose}. However, such approaches are not the preferable choice for industry, always bringing additional consumption.
In our previous study \cite{Liu2020ParametersSE}, we have explored how to build the two-stream backbone network and proposed the hetero-center based triplet loss under the part person feature learning framework.
\textbf{Therefore, to be different from \cite{Liu2020ParametersSE}, in this letter, we try to adopt some simple and typical means to improve the VT-ReID model with only the global person features extracted by the backbone model.}

This letter mainly focuses on the design of an effective baseline from two aspects.
On one hand, some simple and typical means are experimentally explored to obtain the global features, including the pooling and batch normalization operations. On the other hand, the organization manner of sample-based triplet loss and center-based triplet loss are also experimentally explored to guide the network training. To summarize, our dual-granularity triplet loss (DGTL), in a hierarchical fine to coarse granularity manner, could achieve superior performance on RegDB \cite{nguyen2017person} and SYSU-MM01 \cite{wu2017rgb} datasets. It can be a new baseline for VT-ReID with only the global features, through a simple but effective strategy.

\section{Dual-granularity triplet loss based baseline model}
\begin{figure*}
\centering
\includegraphics[width=17cm]{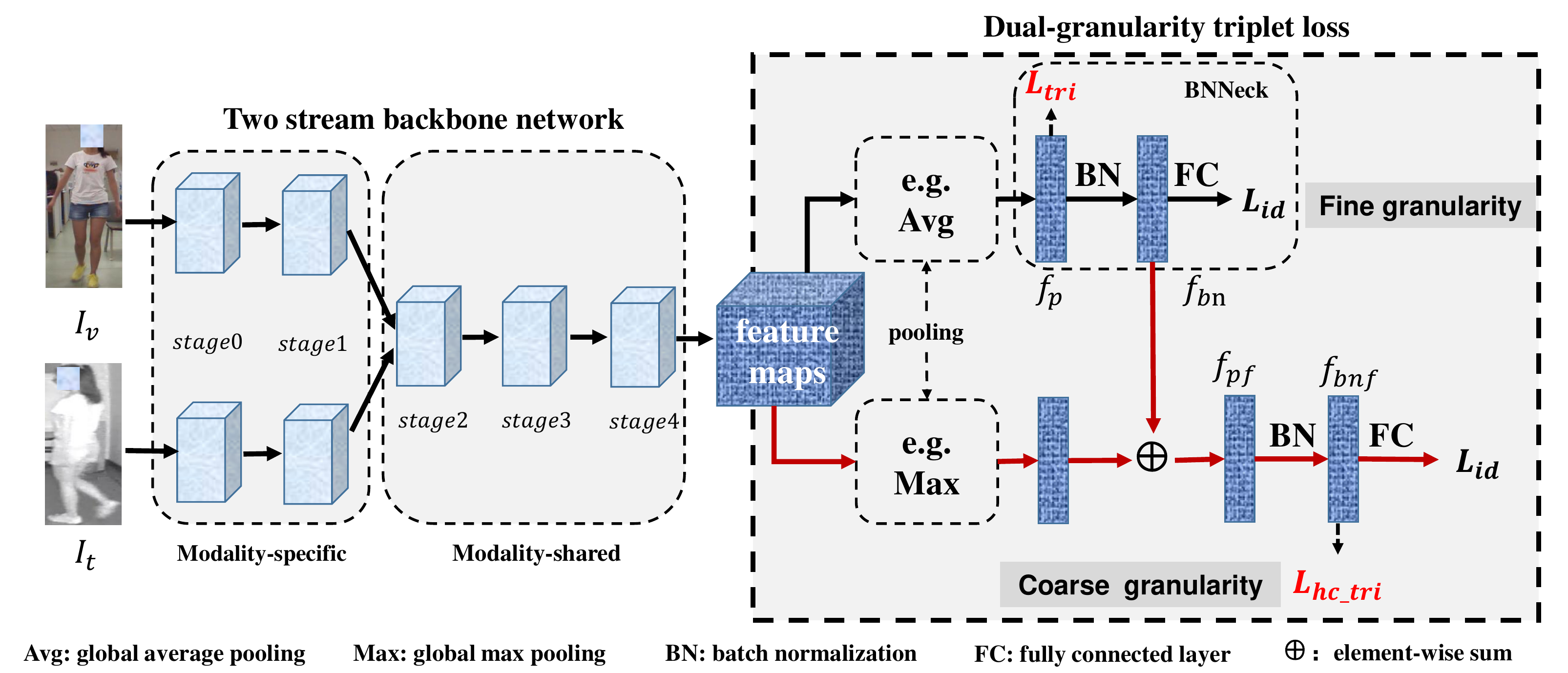}
\caption{The proposed dual-granularity triplet loss based feature learning framework for VT-ReID, including two components: backbone network and dual-granularity triplet loss module. The backbone network contains two modality-specific submodules with independent parameters and one modality-shared submodule with shared parameters.
The dual-granularity triplet loss module focuses on learning high quality person features with inter-class discrimination and intra-class compactness ability.
At the beginning, the 3D feature maps outputted from the backbone are respectively processed by two pooling methods in the fine granularity branch and coarse granularity branch.
In fine granularity branch, the pooling features are supervised by the sample-based triplet loss ($L_{f\_tri}$) and identification loss ($L_{id}$) with BNNeck \cite{Luo2019ASB} to obtain features $f_{bn}$, following the black lines (our baseline). Meanwhile, in the coarse granularity branch, the pooling features are firstly fused with $f_{bn}$. Afterwards, the fused features are supervised by the center-based triplet loss ($L_{c\_tri}$) and identification loss ($L_{id}$) from the coarse granularity level to obtain features $f_{bnf}$, following the additional red lines. During testing, the $f_{bn}$ and $f_{bnf}$  with L2 normalization can be adopted as the person features.}
\label{fig:framework}
\end{figure*}

Fig. \ref{fig:framework} illustrates the framework of our proposed baseline model for VT-ReID, mainly consists of two components: (1) the backbone network, and (2) the dual-granularity triplet loss module.

\subsection{Backbone network}
\label{ssec:backbone}
Based on the observation of our previous study \cite{Liu2020ParametersSE}, we empirically set the backbone with a two-stream network to process images from two different modalities, as shown in Fig. \ref{fig:framework}.
Following \cite{ye2020vipr,Liu2020ParametersSE,ye2020dynamic}, the ResNet50 \cite{he2016deep} model is adopted to construct the backbone network. The shallow convolution block ($layer0$) and the first res-convolution block ($layer1$) are set as the modality-specific submodule with different parameters to learn the modality-specific low-level 3D information from two heterogeneous modalities. Then the remaining 3 res-convolution blocks ($layer2$, $layer3$ and $layer4$) are set as the modality-shared submodule with shared parameters to learn the multi-modality shared mid-level feature representations in a common 3D feature space.

\subsection{Dual-granularity triplet loss module}
Our proposed dual-granularity triplet loss (DGTL) module focuses on three aspects for setting the new VT-ReID baseline, including 1) the pooling methods, 2) the batch normalization neck and 3) loss function.
At each iteration, we adopt the identity-balanced sampling method \cite{ye2020vipr,Liu2020ParametersSE} to construct the mini-batch. For each of $P$ randomly selected person identities, $K$ visible images and $K$ thermal images are randomly sampled, totally including $2*PK$ images.

\subsubsection{Pooling methods}
After obtaining the 3D person feature maps from the backbone network, we should firstly translate them into 1D feature vectors. We experimentally study three kinds of the pooling methods, the global average pooling (Avg), the global max pooling (Max) and the generalized-mean pooling (GeM) \cite{radenovic2018fine}. The results are shown in Tables \ref{tab:ablation_fine_pool} and \ref{tab:ablation_coarse_pool}.\footnote{When the pooling methods in fine and coarse branches are identical (as experimentally set for RegDB dataset), the two branches degenerate to one with a feature skip connection across the batch normalization layer.}

\subsubsection{Batch normalization neck}
The batch normalization neck (BNNeck) \cite{Luo2019ASB} is firstly introduced in VV-ReID to address the inconsistent problem of identification and metric (e.g. triplet) losses in the same embedding space. Namely, the metric loss and identification loss should process different feature vectors, before or after the batch normalization layer. However, in our framework, the BNNeck is only applied in the fine granularity branch, while in the coarse granularity branch the metric and identification losses are both applied to the features after the batch normalization layer ($f_{bnf}$), as shown in Fig. \ref{fig:framework}.

\subsubsection{Dual-granularity triplet loss}
Our previous study \cite{Liu2020ParametersSE} only concentrates on the center-based triplet loss, which is in the coarse granularity level. Here, we simultaneously consider the sample-based triplet loss and center-based triplet loss by constructing two branches, and arrange them in a hierarchical fine to coarse granularity manner, as shown in Fig. \ref{fig:framework}. One branch focuses on the fine granularity level with sample-based triplet loss ($L_{f\_tri}$) and identification loss ($L_{id}$) (our baseline). The other branch processes the fused features, focusing on the coarse granularity level with center-based triplet loss ($L_{c\_tri}$) and identification loss ($L_{id}$).

\textbf{Fine granularity triplet loss}: We equally process person features from the fine granularity level based on each sample whether it is from visible modality or thermal modality. The online hard-mining triplet loss \cite{hermans2017defense} is adopted as our fine granularity triplet loss $L_{f\_tri}$. For each feature $f_a$ in the mini-batch, we can mine the hardest positive $f_p$ and hardest negative $f_n$ to construct the triplet, to compute the fine granularity triplet loss,
\begin{align}\label{eq:loss}
    L_{f\_tri}(f) = \sum\limits_{i=1}^{P} \sum\limits_{a=1}^{2K}
        \Big[m & + \hspace*{-5pt} \max\limits_{p=1 \dots 2K} \hspace*{-5pt} \|f^i_a - f^i_p\|_2  \\
               & - \hspace*{-5pt} \min\limits_{\substack{j=1 \dots P \\ n=1 \dots 2K \\ j \neq i}} \hspace*{-5pt} \| f^i_a - f^j_n \| _2 \Big]_+,\nonumber
\end{align}
where $m$ is the margin, $f_a^i$ denotes the $a^{th}$ image feature of the $i^{th}$ identity in the mini-batch, $[\cdot]_{+} = \max(\cdot, 0)$ represents the standard hinge loss, $\| f_a - f_p \|_2$ denotes the Euclidean distance of two feature vectors $f_a$ and $f_p$.

\textbf{Coarse granularity triplet loss}: We process the visible and thermal person features from the coarse granularity level based on heterogeneous centers of each identity. The hetero-center triplet loss \cite{Liu2020ParametersSE} is adopted as our coarse granularity triplet loss $L_{c\_tri}$. For each identity, we can focus on the only one cross-modality positive center pair and the mined hardest (intra- and inter-modality) negative center pair, to compute the coarse granularity triplet loss,
\begin{small}
\begin{align}\label{eq:loss_ct}
    L_{c\_tri}(f) = & \sum\limits_{i=1}^{P}\Big[mc  +  \|fc^i_v - fc^i_t\|_2  - \min\limits_{\substack{ \texttt{m} \in \{v, t\} \\ j \neq i}}  \| fc^i_v - fc^j_\texttt{m} \| _2 \Big]_+  \\
    & + \sum\limits_{i=1}^{P}\Big[mc  +  \|fc^i_t - fc^i_v\|_2  - \min\limits_{\substack{ \texttt{m} \in \{v, t\} \\ j \neq i}}  \| fc^i_t - fc^j_\texttt{m} \| _2 \Big]_+,\nonumber
\end{align}
\end{small}
where $mc$ is the margin, $fc_{v}^{i} = \frac{1}{K} \sum_{j=1}^{K} f_{v,j}^{i}$, $fc_{t}^{i} = \frac{1}{K} \sum_{j=1}^{K} f_{t,j}^{i}$ are the visible and thermal centers of $i^{th}$ identity, respectively. $f_{v,j}^{i}$ and $f_{t,j}^{i}$ respectively denote the $j^{th}$ visible and thermal image features of $i^{th}$ identity.

\textbf{Dual-granularity triplet loss}: Finally, the overall dual-granularity triplet loss (DGTL) is,
\begin{align}
    L_{all}  =   \overbrace{L_{f\_tri}(f_p) + L_{id}(f_{bn})}^{fine\,\,\,granularity} + \overbrace{L_{c\_tri}(f_{bnf}) + L_{id}(f_{bnf})}^{coarse\,\,\,granularity}, \label{eq:final_loss}
\end{align}
where $f_p$, $f_{bn}$ and $f_{bnf}$ are the global person features as shown in Fig. \ref{fig:framework}. The hierarchical fine to coarse granularity arrangement manner of $L_{f\_tri}$ and $L_{c\_tri}$ is also illustrated in Fig. \ref{fig:framework}. The main contributions of Eq. (\ref{eq:final_loss}) are 1) the organization of $L_{f\_tri}$ and $L_{c\_tri}$, 2) the processing features for each triplet loss, corresponding to the position of BNNeck.

\section{Experiments}
We evaluate the effectiveness of our proposed method for VT-ReID on two public datasets, RegDB \cite{nguyen2017person} and  SYSU-MM01 \cite{wu2017rgb}.
The implementation\footnote{\url{https://github.com/hijune6/DGTL-for-VT-ReID}} of our method is with the Pytorch framework. The training and testing procedures are following the official settings as done in \cite{Ye2020DeepLF,Liu2020ParametersSE}.
For the $PK$ sampling strategy, we set $P = 8$, $K=4$ for the RegDB, and $P=6$, $K=8$ for the SYSU-MM01.
The pooling method is Max in both fine and coarse branches for RegDB, while Avg in fine branch and Max in coarse branch for SYSU-MM01.
We set $m = 0.3$, $mc = 0.3$ for RegDB, and $mc = 0.8$ for SYSU-MM01.
The fusion method is element-wise sum.

\subsection{Comparison to the state-of-the-art}

\begin{table}
\caption{Comparison to the state-of-the-art methods on the RegDB datasets. Re-identification rates at rank1 and mAP (\%).}
\label{tab:sota_regdb}
  \centering
  \begin{tabular}{lc|c|c|c||c|c}
    \toprule[2pt]
    \multicolumn{3}{c|}{}  & \multicolumn{2}{c||}{\emph{Visible to Thermal}} & \multicolumn{2}{c}{\emph{Thermal to Visible}} \\ \hline
      \multicolumn{2}{c|}{Methods} & Venue &   rank1   & mAP   &  rank1  & mAP      \\ \toprule[1pt]
      \multicolumn{2}{c|}{CMSP \cite{wu2020rgb}} & IJCV20 & 65.07   & 64.50  & -   & -  \\
      \multicolumn{2}{c|}{HAT \cite{ye2020vipr}} & TIFS20 & 71.83  & 67.56 & 70.02  & 66.30  \\
      \multicolumn{2}{c|}{MSR \cite{Feng2020LearningMR}} & TIP20 & 48.43  & 48.67  & -  & -  \\
      \multicolumn{2}{c|}{MACE \cite{Ye2020CrossModalityPR}} & TIP20 & 72.37 & 69.09 & 72.12 & 68.57 \\
      \multicolumn{2}{c|}{Hi-CMD \cite{choi2020hi}} & CVPR20 & 70.93  & 66.04   &  - & -   \\
      \multicolumn{2}{c|}{CML \cite{Ling2020ClassAwareMM}} & MM20 & 59.81 & 60.86  &  - & -   \\
      \multicolumn{2}{c|}{JSIA \cite{Wang2020CrossModalityPG}} & AAAI20 & 48.10  & 48.90 & 48.50  & 49.30 \\
      \multicolumn{2}{c|}{XIV \cite{Li2020InfraredVisibleCP}} & AAAI20 & 62.21  & 60.18  &  - & -   \\
      \multicolumn{2}{c|}{DDAG \cite{ye2020dynamic}} & ECCV20 & 69.34  & 63.46 & 68.06  & 61.08 \\  \hline
      \multirow{2}{*}{DGTL} & $f_{bn}$& \multirow{2}{*}{ours} & 83.56 & 73.36 & 81.27 & 71.22 \\
                            & $f_{bnf}$& & \textbf{83.92} & \textbf{73.78} & \textbf{81.59} & \textbf{71.65} \\ \hline \hline
      \multicolumn{2}{c|}{HcTri \cite{Liu2020ParametersSE}} & TMM20 & 91.05 & 83.28 & 89.30 & 81.46 \\
      \toprule[2pt]
  \end{tabular}
\end{table}

\begin{table}
\caption{Comparison to the state-of-the-art methods on the SYSU-MM01 datasets. Re-identification rates at rank1 and mAP (\%).}
\label{tab:sota_sysu}
  \centering
  \begin{tabular}{lc|c|c|c||c|c}
    \toprule[2pt]
    \multicolumn{3}{c|}{}  & \multicolumn{2}{c||}{\emph{All search}} & \multicolumn{2}{c}{\emph{Indoor search}} \\ \hline
      \multicolumn{2}{c|}{Methods} & Venue &   rank1   & mAP   &  rank1  & mAP      \\ \toprule[1pt]
      \multicolumn{2}{c|}{CMSP \cite{wu2020rgb}} & IJCV20 & 43.56   & 44.98  & 48.62   & 57.50  \\
      \multicolumn{2}{c|}{HAT \cite{ye2020vipr}} & TIFS20 & 55.29  & 53.89 & 62.10  & 69.37  \\
      \multicolumn{2}{c|}{MSR \cite{Feng2020LearningMR}} & TIP20 & 37.35  & 38.11  & 39.64  & 50.88  \\
      \multicolumn{2}{c|}{MACE \cite{Ye2020CrossModalityPR}} & TIP20 & 51.64 & 50.11 & 57.35 & 64.79 \\
      \multicolumn{2}{c|}{Hi-CMD \cite{choi2020hi}} & CVPR20 & 34.94  & 35.94   &  - & -   \\
      \multicolumn{2}{c|}{CML \cite{Ling2020ClassAwareMM}} & MM20 & 51.80 & 51.21 & 54.98 & 63.7 \\
      \multicolumn{2}{c|}{JSIA \cite{Wang2020CrossModalityPG}} & AAAI20 & 38.10  & 36.90 & 43.80  & 52.90 \\
      \multicolumn{2}{c|}{XIV \cite{Li2020InfraredVisibleCP}} & AAAI20 & 49.92  & 50.73  &  - & -   \\
      \multicolumn{2}{c|}{DDAG \cite{ye2020dynamic}} & ECCV20 & 54.75  & 53.02 & 61.02  & 67.98 \\  \hline
      \multirow{2}{*}{DGTL} & $f_{bn}$ & \multirow{2}{*}{ours} & 54.66 & 52.72 & 59.21 & 66.27 \\
                                & $f_{bnf}$& & \textbf{57.34} & \textbf{55.13} & \textbf{63.11} & \textbf{69.20} \\ \hline  \hline
      \multicolumn{2}{c|}{HcTri \cite{Liu2020ParametersSE}} & TMM20 & 61.68 & 57.51 & 63.41 & 68.17 \\
      \toprule[2pt]
  \end{tabular}
\end{table}

\begin{figure*}
\centering
\begin{tabular}{c@{\hspace{2mm}}c}
\includegraphics[width=8.9cm]{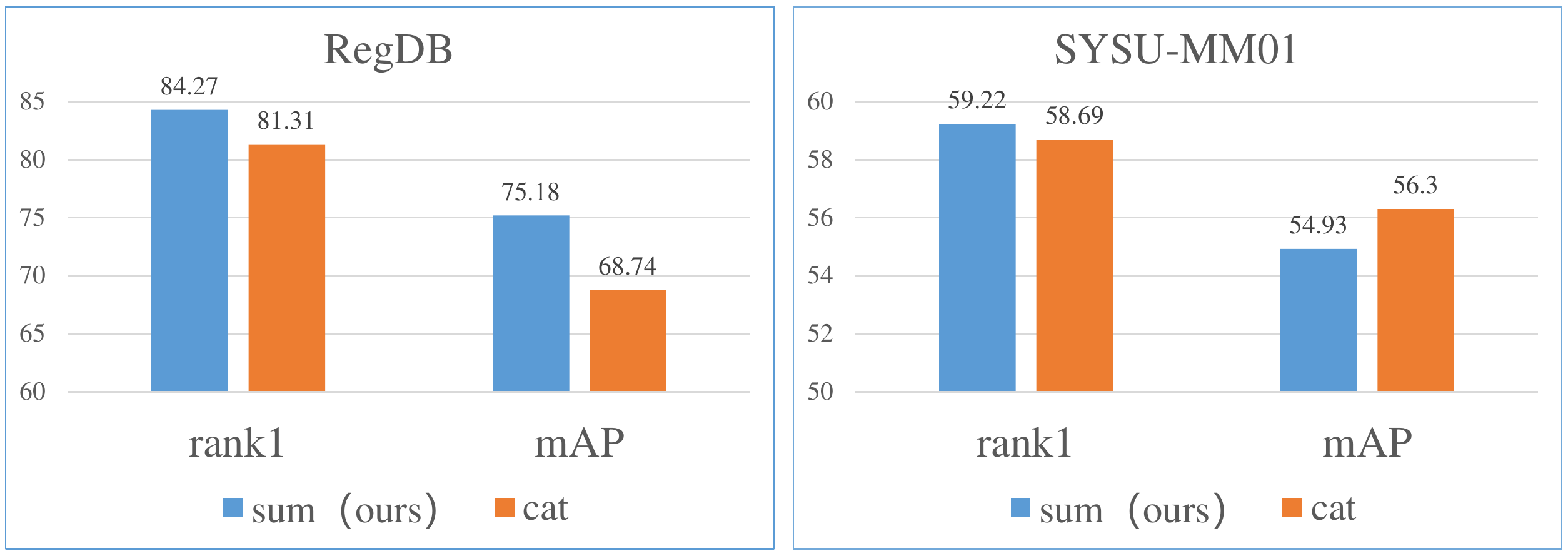} &
\includegraphics[width=8.9cm]{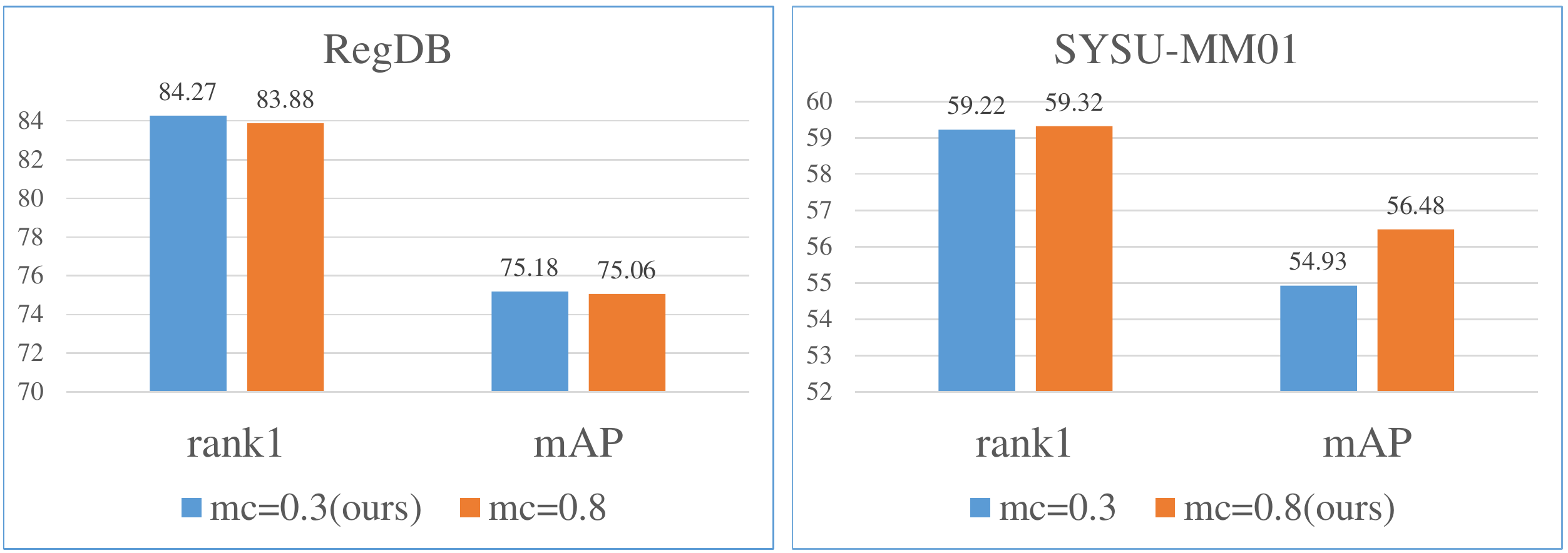} \\ (a) Fusion methods  & (b) Margin parameter $mc$ in $L_{c\_tri}$
\end{tabular}
\caption{The effects of (a) fusion methods (sum: element-wise sum, cat: concatenation) and (b) margin parameter $mc$ in $L_{c\_tri}$ on RegDB and SYSU-MM01 datasets. Re-identification rates of rank1 and mAP (\%).}
\label{fig:fuse_margin}
\end{figure*}

In this section, our DGTL with only the global features is compared to some state-of-the-art VT-ReID methods, recently published in 2020. The results on the RegDB and SYSU-MM01 datasets are listed in Tables \ref{tab:sota_regdb} and \ref{tab:sota_sysu}, respectively. In this subsection, the mean results of 10 trials are reported following the standard dataset settings.
They show that our proposed DGTL method can achieve much better performance, especially compared to those methods with only the global features (CMSP \cite{wu2020rgb},HAT \cite{ye2020vipr},MSR \cite{Feng2020LearningMR},MACE \cite{Ye2020CrossModalityPR},Hi-CMD \cite{choi2020hi},CML \cite{Ling2020ClassAwareMM},JSIA \cite{Wang2020CrossModalityPG} and XIV \cite{Li2020InfraredVisibleCP}), even outperforming the DDAG \cite{ye2020dynamic} method, which adopts the part-aggregated feature learning to refine the person features.
Moreover, the results based on the $f_{bn}$ feature, just the direct output of the ResNet50 model, also can achieve satisfactory performance, even similar to those results based on $f_{bnf}$ feature on RegDB dataset. It demonstrates the effectiveness of our dual-granularity triplet loss module with a simple but effective strategy, which truly can be a strong VT-ReID baseline to  boost future research with high quality.

Our DGTL method performs worse than HcTri \cite{Liu2020ParametersSE}, which is our previous study for part feature learning with much more model parameters and training tricks.

\subsection{Alation experiments}
We evaluate the effectiveness of our proposed DGTL module, including three components, pooling methods, loss functions organization and the BNNeck configuration. To simply show the effectiveness of different components, during the ablation experiments, only one trial experimental results are reported, rather than the mean results of 10 trials.

\begin{table}
\scriptsize
\caption{The effects of different pooling methods in baseline network (the fine granularity branch). Re-identification rates of rank1 and mAP (\%).}
\label{tab:ablation_fine_pool}
  \centering
  \setlength{\tabcolsep}{0.5cm}
  \begin{tabular}{c|c|c|c|c}
   \toprule[2pt]
     & \multicolumn{2}{c|}{RegDB} & \multicolumn{2}{c}{SYSU-MM01}\\ \hline
    Fine & rank1 & mAP  & rank1 & mAP  \\ \toprule[1pt]
    Avg & 70.63 & 65.13 & \textbf{58.66} & \textbf{54.57} \\
    Max & \textbf{80.49} & \textbf{71.23} & 49.04 & 47.35  \\
    GeM & 75.39 & 67.72 & 55.85 & 53.25  \\
     \toprule[2pt]
  \end{tabular}
\end{table}

\begin{table}
\scriptsize
\caption{The effects of different pooling methods in the coarse granularity branch. The pooling method in fine granularity branch is Max for RegDB dataset and Avg for SYSU-MM01 dataset.  Re-identification rates of rank1 and mAP (\%).}
\label{tab:ablation_coarse_pool}
  \centering
  \setlength{\tabcolsep}{0.5cm}
  \begin{tabular}{c|c|c|c|c}
   \toprule[2pt]
      \multicolumn{5}{c}{RegDB} \\ \hline
    Fine & Coarse & features & rank1 & mAP    \\ \toprule[1pt]
   \multirow{6}{*}{Max} & \multirow{2}{*}{Avg} & $f_{bn}$ & 83.59 & 72.08 \\
                       &  & $f_{bnf}$ & 83.30 & 72.13  \\ \cline{2-5}
    &\multirow{2}{*}{Max} & $f_{bn}$ & 83.64 & 74.53  \\
                        & & $f_{bnf}$ & \textbf{84.27} & \textbf{75.18}  \\ \cline{2-5}
    &\multirow{2}{*}{GeM} & $f_{bn}$ & 77.82 & 68.59  \\
                        & & $f_{bnf}$ & 77.28 & 68.17  \\ \toprule[1pt] \toprule[1pt]
       \multicolumn{5}{c}{SYSU-MM01} \\ \hline
   \multirow{6}{*}{Avg} & \multirow{2}{*}{Avg} & $f_{bn}$ & 56.59 & 54.02 \\
                       &  & $f_{bnf}$ &  56.74 & 54.23 \\ \cline{2-5}
    &\multirow{2}{*}{Max} & $f_{bn}$ & 55.35 & 52.44  \\
                        & & $f_{bnf}$ & \textbf{59.22} &  54.93 \\ \cline{2-5}
    &\multirow{2}{*}{GeM} & $f_{bn}$ & 56.22 & 53.98  \\
                        & & $f_{bnf}$ & 57.67 & \textbf{55.14}  \\
     \toprule[2pt]
  \end{tabular}
\end{table}

Tables \ref{tab:ablation_fine_pool} and \ref{tab:ablation_coarse_pool} list the results of different pooling method arrangements in fine and coarse granularity branches, respectively. Different pooling methods truly perform differently, always with large gaps (e.g. Avg vs. Max: 70.63 vs. 80.49, rank1 on regdb dataset). Therefore, the pooling method is a key factor for constructing the VT-ReID baseline.

Table \ref{tab:ablation_loss_org} lists the results of different triplet loss arrangements in the fine and coarse granularity branches. In our baseline methods (only with the fine granularity branch), the combination of $L_{f\_tri}$ and $L_{c\_tri}$ truly can improve the VT-ReID performance. As to the dual-granularity setting, the arrangements of $L_{f\_tri}$ and $L_{c\_tri}$ have impact on the performance. In summary, our proposed hierarchical fine to coarse (f2c) granularity manner could obtain the best performance.

\begin{table}
 \tiny
\caption{The ablation study of triplet losses organizing in the fine and coarse granularity branches, center-based triplet loss ($L_{c\_tri}$) and sample-based triplet loss ($L_{f\_tri}$). Re-identification rates of rank1 and mAP (\%).}
\label{tab:ablation_loss_org}
  \centering
  \begin{tabular}{c|c|c|c|c|c|c|c}
   \toprule[2pt]
    \multicolumn{4}{c|}{} & \multicolumn{2}{c|}{RegDB} & \multicolumn{2}{c}{SYSU-MM01}\\ \hline
    index& Fine & Coarse & features & rank1 & mAP & rank1 & mAP \\ \toprule[1pt]
    \ding{172}&  $L_{f\_tri}$ & $\times$ & $f_{bn}$ &  80.49 & 71.23 & 58.66 & 54.57  \\
    \ding{173}&  $L_{c\_tri}$ & $\times$ & $f_{bn}$ & 80.58  & 65.57 & 53.22 & 50.34   \\
    \ding{174}&  $L_{f\_tri}+L_{c\_tri}$ & $\times$ & $f_{bn}$ & 83.74  & 74.81 & 57.95 & 55.48   \\ \hline
    \multirow{2}{*}{f2f} & \multirow{2}{*}{$L_{f\_tri}$} & \multirow{2}{*}{$L_{f\_tri}$} & $f_{bn}$ & 80.78 & 72.09 & 53.75 & 50.45  \\
                          &          &                             & $f_{bnf}$& 78.16 & 71.36 & 57.51 & 54.70 \\ \hline
    \multirow{2}{*}{c2c} & \multirow{2}{*}{$L_{c\_tri}$} & \multirow{2}{*}{$L_{c\_tri}$} & $f_{bn}$ & 81.70 & 66.65 & 54.72 & 50.64  \\
                        &            &                             & $f_{bnf}$ & 82.33 & 66.68 & 57.45 & 53.44  \\ \hline
    \multirow{2}{*}{c2f} & \multirow{2}{*}{$L_{c\_tri}$} & \multirow{2}{*}{$L_{f\_tri}$} & $f_{bn}$ & 80.53  & 72.90 & 53.88 & 51.16   \\
                        &            &                             & $f_{bnf}$& 78.74  & 72.58 & 57.53 & 55.82 \\ \hline
    \multirow{2}{*}{f2c} & \multirow{2}{*}{$L_{f\_tri}$} & \multirow{2}{*}{$L_{c\_tri}$} & $f_{bn}$ & 83.64 & 74.53  & 55.96 & 53.85  \\
                        &            &                             & $f_{bnf}$  & \textbf{84.27}  & \textbf{75.18} & \textbf{59.32} & \textbf{56.48}  \\
     \toprule[2pt]
  \end{tabular}
\end{table}

\begin{table}
\scriptsize
\caption{The ablation study of BNNeck \cite{Luo2019ASB} in the fine and coarse granularity branches. Whether applying the BNNeck or not, i.e., where the triplet loss should be adopted, for the pool features ($f_p$,$f_{pf}$)  or the batch normalization features ($f_bn$, $f_{bnf}$)? Re-identification rates of rank1 and mAP (\%).}
\label{tab:ablation_bnneck}
  \centering
  \begin{tabular}{c|c|c|c|c|c}
   \toprule[2pt]
    Fine & Coarse & \multicolumn{2}{c|}{RegDB} & \multicolumn{2}{c}{SYSU-MM01}\\ \hline
     $L_{f\_tri}$ & $L_{c\_tri}$ &  rank1 & mAP & rank1 & mAP \\ \toprule[1pt]
     $f_p$ & $\times$ &  80.49  & 71.23 & 58.66 & 54.57  \\
     $f_{bn}$ & $\times$ &  77.43  & 70.53 & 49.78 & 48.99   \\
     $f_p$ & $f_{pf}$ & 75.97  & 66.52 & 58.37 & 54.49     \\
     $f_{bn}$ & $f_{bnf}$ & 81.17  & 72.57 & 54.77 & 52.42  \\
     $f_p$ & $f_{bnf}$ & \textbf{84.27}  & \textbf{75.18} & \textbf{59.32} &  \textbf{56.48}   \\
     \toprule[2pt]
  \end{tabular}
\end{table}

Table \ref{tab:ablation_bnneck} shows that the BNNeck \cite{Luo2019ASB} module only applied in the fine granularity branch is the best setting.
Moreover, Fig. \ref{fig:fuse_margin} also illustrates the effects of different fusion methods and the margin parameter in $L_{c\_tri}$.

The best performances for two datasets are with different configurations. The reason may lie in the image conditions. For RegDB, the visible and corresponding thermal images are well aligned. While for SYSU-MM01, the visible and corresponding infrared images are with arbitrary poses and views.

\section{Conclusions}
In this letter, we propose a strong VT-ReID baseline with a simple but effective strategy. To our best knowledge, it can achieve the best performance with only the global features extracted by the backbone model.
Our proposed DGTL method arranges the sample-based triplet loss and center-based triplet loss in a hierarchical fine to coarse granularity manner. Some simple configurations of typical operations, e.g. the pooling methods and batch normalization, are also explored for VT-ReID tasks.
We hope that this study can promote the VT-ReID research with high quality.


\bibliographystyle{IEEEtrans}
\bibliography{reid}

\begin{thebibliography}{10}
\providecommand{\url}[1]{#1}
\csname url@samestyle\endcsname
\providecommand{\newblock}{\relax}
\providecommand{\bibinfo}[2]{#2}
\providecommand{\BIBentrySTDinterwordspacing}{\spaceskip=0pt\relax}
\providecommand{\BIBentryALTinterwordstretchfactor}{4}
\providecommand{\BIBentryALTinterwordspacing}{\spaceskip=\fontdimen2\font plus
\BIBentryALTinterwordstretchfactor\fontdimen3\font minus
  \fontdimen4\font\relax}
\providecommand{\BIBforeignlanguage}[2]{{%
\expandafter\ifx\csname l@#1\endcsname\relax
\typeout{** WARNING: IEEEtranS.bst: No hyphenation pattern has been}%
\typeout{** loaded for the language `#1'. Using the pattern for}%
\typeout{** the default language instead.}%
\else
\language=\csname l@#1\endcsname
\fi
#2}}
\providecommand{\BIBdecl}{\relax}
\BIBdecl

\bibitem{choi2020hi}
S.~Choi, S.~Lee, Y.~Kim, T.~Kim, and C.~Kim, ``Hi-cmd: Hierarchical
  cross-modality disentanglement for visible-infrared person
  re-identification,'' in \emph{CVPR}, 2020, pp. 10\,257--10\,266.

\bibitem{Feng2020LearningMR}
Z.~Feng, J.~Lai, and X.~Xie, ``Learning modality-specific representations for
  visible-infrared person re-identification,'' \emph{IEEE TIP}, vol.~29, pp.
  579--590, 2020.

\bibitem{he2016deep}
K.~He, X.~Zhang, S.~Ren, and J.~Sun, ``Deep residual learning for image
  recognition,'' in \emph{CVPR}, 2016, pp. 770--778.

\bibitem{Li2020InfraredVisibleCP}
D.~Li, X.~Wei, X.~Hong, and Y.~Gong, ``Infrared-visible cross-modal person
  re-identification with an x modality,'' in \emph{AAAI}, 2020, pp. 4610--4617.

\bibitem{Ling2020ClassAwareMM}
Y.~Ling, Z.~Zhong, Z.~Luo, P.~Rota, S.~Li, and N.~Sebe, ``Class-aware modality
  mix and center-guided metric learning for visible-thermal person
  re-identification,'' in \emph{ACM MM}, 2020, pp. 889--897.

\bibitem{Liu2019Gallery}
H.~Liu and J.~Cheng, ``Gallery based k-reciprocal-like re-ranking for heavy
  cross-camera discrepancy in person re-identification,''
  \emph{Neurocomputing}, vol. 333, pp. 64--75, 2019.

\bibitem{liu2020enhancing}
H.~Liu, J.~Cheng, W.~Wang, Y.~Su, and H.~Bai, ``Enhancing the discriminative
  feature learning for visible-thermal cross-modality person
  re-identification,'' \emph{Neurocomputing}, vol. 398, pp. 11--19, 2020.

\bibitem{Liu2020ParametersSE}
H.~Liu, X.~Tan, and X.~Zhou, ``Parameter sharing exploration and hetero-center
  triplet loss for visible-thermal person re-identification,'' \emph{IEEE TMM},
  pp. 1--1, 2020.

\bibitem{Luo2019ASB}
H.~Luo, W.~Jiang, Y.~Gu, F.~Liu, X.~Liao, S.~Lai, and J.~Gu, ``A strong
  baseline and batch normalization neck for deep person re-identification,''
  \emph{IEEE TMM}, vol.~22, no.~10, pp. 2597--2609, 2020.

\bibitem{nguyen2017person}
D.~Nguyen, H.~Hong, K.~Kim, and K.~Park, ``Person recognition system based on a
  combination of body images from visible light and thermal cameras,''
  \emph{Sensors}, vol.~17, no.~3, p. 605, 2017.

\bibitem{radenovic2018fine}
F.~Radenovi{\'c}, G.~Tolias, and O.~Chum, ``Fine-tuning cnn image retrieval
  with no human annotation,'' \emph{IEEE TPAMI}, vol.~41, no.~7, pp.
  1655--1668, 2018.

\bibitem{su2017pose}
C.~Su, J.~Li, S.~Zhang, J.~Xing, W.~Gao, and Q.~Tian, ``Pose-driven deep
  convolutional model for person re-identification,'' in \emph{ICCV}, 2017, pp.
  3980--3989.

\bibitem{Wang2020CrossModalityPG}
G.~Wang, T.~Zhang, Y.~Yang, J.~Cheng, J.~Chang, X.~Liang, and Z.~Hou,
  ``Cross-modality paired-images generation for rgb-infrared person
  re-identification,'' in \emph{AAAI}, 2020, pp. 12\,144--12\,151.

\bibitem{Wang2020MPMN}
P.~{Wang}, Z.~{Zhao}, F.~{Su}, Y.~{Zhao}, H.~{Wang}, L.~{Yang}, and Y.~{Li},
  ``Deep multi-patch matching network for visible thermal person
  re-identification,'' \emph{IEEE TMM}, pp. 1--1, 2020.

\bibitem{wu2020rgb}
A.~Wu, W.-S. Zheng, S.~Gong, and J.~Lai, ``Rgb-ir person re-identification by
  cross-modality similarity preservation,'' \emph{IJCV}, vol. 128, pp.
  1765--1785, 2020.

\bibitem{wu2017rgb}
A.~Wu, W.~Zheng, H.~Yu, S.~Gong, and J.~Lai, ``Rgb-infrared cross-modality
  person re-identification,'' in \emph{ICCV}, 2017, pp. 5380--5389.

\bibitem{Ye2020CrossModalityPR}
M.~Ye, X.~Lan, and Q.~Leng, ``Cross-modality person re-identification via
  modality-aware collaborative ensemble learning,'' \emph{IEEE TIP}, vol.~29,
  pp. 9387--9399, 2020.

\bibitem{tifs19vtreid}
M.~Ye, X.~Lan, Z.~Wang, and P.~C. Yuen, ``Bi-directional center-constrained
  top-ranking for visible thermal person re-identification,'' \emph{IEEE TIFS},
  vol.~15, pp. 407--419, 2020.

\bibitem{ye2020dynamic}
M.~Ye, J.~Shen, D.~J. Crandall, L.~Shao, and J.~Luo, ``Dynamic dual-attentive
  aggregation learning for visible-infrared person re-identification,'' in
  \emph{ECCV}, 2020.

\bibitem{Ye2020DeepLF}
M.~Ye, J.~Shen, G.~Lin, T.~Xiang, L.~Shao, and S.~C.~H. Hoi, ``Deep learning
  for person re-identification: A survey and outlook,'' \emph{arXiv preprint
  arXiv:2001.04193}, 2020.

\bibitem{ye2020vipr}
M.~Ye, J.~Shen, and L.~Shao, ``Visible-infrared person re-identification via
  homogeneous augmented tri-modal learning,'' \emph{IEEE TIFS}, vol.~16, pp.
  728--739, 2021.

\end{thebibliography}
\end{document}